\documentclass{article}
\usepackage{arxiv}

\usepackage{amsmath}
\usepackage{amssymb}
\usepackage{amsfonts}
\usepackage{booktabs}
\usepackage{multirow}
\usepackage{hyperref}
\usepackage{enumitem}
\usepackage{microtype}
\usepackage{siunitx}

\usepackage{graphicx} 

\title{Neural networks for Text-to-Speech evaluation}

\author{
  \textbf{Ilya Trofimenko$^1$, David Kocharyan$^1$, Aleksandr Zaitsev$^1$,} \\
  \textbf{Pavel Repnikov$^1$, Mark Levin$^2$, Nikita Shevtsov$^3$} \\[1.5ex]
  \normalfont $^1$ HSE University, Moscow\\
  \normalfont $^2$ HSE University, Saint-Petersburg\\
  \normalfont $^3$ Institute for System Programming, Russian Academy of Sciences\\[1.5ex]
  \normalfont \texttt{troila024@gmail.com, drkocharyan67@gmail.com, AlZayts@yandex.ru,} \\
  \normalfont \texttt{rtfiof@gmail.com, levinmark907@gmail.com}
}

\begin{document}

\maketitle

\begin{abstract}
Ensuring that Text-to-Speech (TTS) systems deliver human-perceived quality at scale is a central challenge for modern speech technologies. Human subjective evaluation protocols such as Mean Opinion Score (MOS) and Side-by-Side (SBS) comparisons remain the de facto gold standards, yet they are expensive, slow, and sensitive to pervasive assessor biases. This study addresses these barriers by formulating, and implementing a suite of novel neural models designed to approximate expert judgments in both relative (SBS) and absolute (MOS) settings. For relative assessment, we propose NeuralSBS, a HuBERT-backed model achieving 73.7\% accuracy (on SOMOS dataset). For absolute assessment, we introduce enhancements to MOSNet using custom sequence-length batching, as well as WhisperBert, a multimodal stacking ensemble that combines Whisper audio features and BERT textual embeddings via weak learners. Our best MOS models achieve a Root Mean Square Error (RMSE) of $\sim$0.40, significantly outperforming the human inter-rater RMSE baseline of 0.62. Furthermore, our ablation studies reveal that naively fusing text via cross-attention can degrade performance, highlighting the effectiveness of ensemble-based stacking over direct latent fusion. We additionally report negative results with SpeechLM-based architectures and zero-shot LLM evaluators (Qwen2-Audio, Gemini 2.5 flash preview), reinforcing the necessity of dedicated metric learning frameworks.
\end{abstract}


\section{Introduction}

Modern Text-to-Speech (TTS) systems have transitioned from concatenative methods to neural architectures capable of producing high-fidelity, natural-sounding speech, enabling widespread deployment in voice assistants, accessibility technology, and dynamic content generation. As model capacity has grown, the bottleneck for iteration has shifted from synthesis to evaluation: measuring perceived naturalness, prosody, intelligibility, and absence of artifacts. Human subjective protocols such as Mean Opinion Score (MOS) and Side-by-Side (SBS) comparisons remain indispensable due to their direct grounding in perception, but they suffer from time and cost inefficiency, inter- and intra-rater variability, and challenges in maintaining consistency across large-scale experiments. For instance, robust evaluation of just two models often requires thousands of generated clips and a costly, time-consuming dependency on a large pool of human assessors for manual annotation, which severely bottlenecks rapid regression testing in CI/CD pipelines.

The current technical landscape reveals a critical gap: classical objective measures operate at low cost but only weakly correlate with human perception of naturalness, while human evaluation is too slow and expensive. This gap motivates the creation of specialized, neural evaluators grounded in modern self-supervised audio representations. 

The object of this research is the automated evaluation of synthesized speech, focusing on designing neural metrics that predict both relative preference (SBS) and absolute scores (MOS) with high alignment to human judgments. To achieve this, we introduce a collaborative-filtering-inspired method to mitigate rater bias via data standardization (Std SOMOS); we implement and refine the NeuralSBS architecture for pairwise comparison; we optimize absolute MOS predictors (MOSNet); and we propose a novel multimodal ensemble architecture, WhisperBert. By demonstrating that our models can achieve an RMSE of $\sim$0.40 (compared to the human baseline of 0.62) and an SBS accuracy of $\sim$74\% (comparable to human subjective agreement), this work aims to deliver practically useful models for automated TTS quality control.

\section{Literature review}

Evaluation of synthesized speech quality spans two principal paradigms: subjective and objective. metrics Subjective protocols --- most notably MOS panels and SBS preference tests --- are considered the gold standard. However, both are vulnerable to rater-scale heterogeneity. In contrast, traditional objective measures (e.g., PESQ, STOI, Mel-cepstral distortion) correlate only weakly with perceived naturalness. 

The evolution of automatic TTS evaluation has closely tracked advances in deep learning, with the vast majority of research focusing on absolute quality prediction. At the forefront of this domain is UTMOS \cite{utmos}, the top-performing ensemble system from the VoiceMOS Challenge 2022 \cite{voicemos}, which firmly established that fine-tuning large-scale Self-Supervised Learning (SSL) representations (e.g., wav2vec 2.0, HuBERT \cite{hubert}) yields state-of-the-art correlation with human raters. This marks a dramatic evolution from early breakthroughs, such as MOSNet \cite{mosnet}, which relied on convolutional feature extractors coupled with temporal modeling on spectrograms. The development and robust training of these modern SSL-based evaluators have been heavily catalyzed by pivotal resources like the SOMOS dataset \cite{somos}, which provides over 20,000 clips synthesized by 200 systems alongside 360,000 MOS ratings.

As the field expanded, specialized absolute-quality models emerged to address specific bottlenecks. For instance, DNSMOS \cite{dnsmos} was developed as a robust, non-intrusive metric to evaluate noisy and enhanced speech, trained on massive datasets to operate without a clean reference signal. Concurrently, the necessity for rapid evaluation in CI/CD pipelines drove the creation of lightweight architectures like DistilMOS \cite{distilmos}, which distill the knowledge of cumbersome SSL models into faster, deployment-friendly networks without sacrificing human correlation. Other notable efforts include NISQA \cite{nisqa} for predicting network-degraded speech quality.

Despite this massive progress in absolute MOS prediction, automated pairwise preference evaluating remains severely underexplored in speech synthesis. The Neural Side-by-Side (NSBS) \cite{nsbs} framework for image evaluation task offers a foundation by imposing an antisymmetric bilinear pooling layer to encode the logical identity 
\(P(A > B) = 1 - P(B > A)\). Coupling this inductive bias with SSL image encoders has proven highly effective. Crucially, while automated pairwise preference models have become an useful tool in computer vision (such as LPIPS \cite{lpips} for automated A/B generative image assessment), a dedicated, robust SBS predictor for synthesized speech has been apparently absent. Therefore, translating these relative preference architectures from the visual domain to speech represents a new step toward fully automated, human-aligned TTS evaluation.

While Large Language Models (LLMs) with native audio processing capabilities (e.g., SpeechGPT \cite{speechgpt}) suggest an end-to-end route to evaluation agents, calibration and fine-grained score prediction without dense supervision remain challenging. Our preliminary experiments with zero-shot LLMs for MOS and SBS scoring yielded sub-optimal performance compared to specialized architectures, reinforcing the need for dedicated metric learning frameworks.

\section{Methodology}

\subsection{Data standardization and preprocessing}

A principal source of label noise in human evaluation is rater-specific scale usage. Drawing inspiration from collaborative filtering, we compute within-rater standardized scores to mitigate bias. Let \(y_{r,i}\) be the raw score assigned by rater \(r\) to item \(i\). We compute the standardized score \(z_{r,i}=(y_{r,i}-\mu_r)/\sigma_r\), where \(\mu_r\) and \(\sigma_r\) are the mean and standard deviation of rater \(r\)'s scores. These are then affinely rescaled to the \([1,5]\) range, yielding the \textbf{Std SOMOS} targets. This transformation equalizes scale usage, increases effective resolution by introducing continuous scores between the original discrete values (e.g., 3.03 or 3.5), and empirically tightens the relationship between features and labels.

We use the SOMOS dataset \cite{somos}, generating 90,000 SBS pairs from the 20,000 English clips. The data is split 70/30 into training and testing sets, ensuring no text leakage occurs across splits.

\subsection{Data augmentation}
\label{sec:augmentation}

To increase the diversity of the training set, we apply a two-stage augmentation pipeline. In the first stage, we apply signal-level perturbations: additive white and pink noise injection, artificial micro-gap insertion (simulating brief audio dropouts), and frequency response modification. In the second stage, we perform phonetically motivated transformations: duration modification of phoneme segments, room-impulse-response convolution (reverberation), systematic voiced/unvoiced consonant substitution, and pitch correction.

\subsection{Architectures for relative evaluation (SBS)}

We pose the SBS metric learning problem as learning a function \(g:(x_a, x_b, t)\mapsto \hat{p}\in[0,1]\) representing the probability that audio clip \(x_a\) is preferred over \(x_b\). 

\textbf{NeuralSBS:} Following \cite{nsbs}, we enforce antisymmetry by constructing a score \(s\) from utterance-level embeddings \(z_a, z_b\in\mathbb{R}^d\) extracted via a HuBERT encoder (12 layers, \SI{960}{\hour} English speech):
\begin{align}
  s &= z_a^\top W z_b - z_b^\top W z_a \\
  \hat{p} &= \sigma(s)
\end{align}
where \(W \in \mathbb{R}^{d \times d}\) is a learnable matrix. This ensures \(g(x_a,x_b) = 1 - g(x_b,x_a)\), preventing the classifier from learning asymmetric shortcuts.

\begin{figure}[th]
    \centering
    \includegraphics[width=0.8\linewidth]{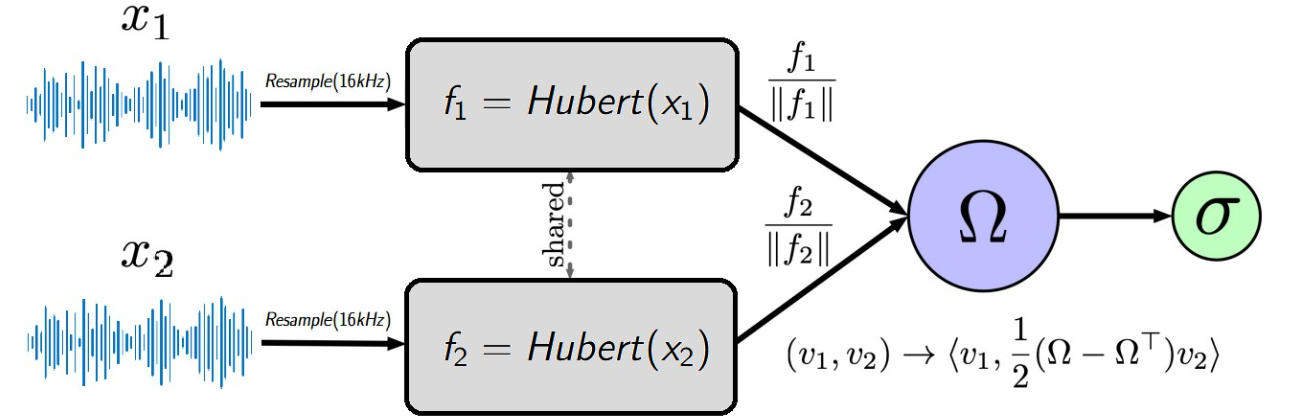}
    \caption{The NeuralSBS architecture. Audio clips A and B are encoded by a shared HuBERT model, followed by temporal averaging to produce utterance embeddings. These embeddings are processed by an antisymmetric bilinear layer to produce a preference score.}
    \label{fig:neuralsbs}
\end{figure}

\begin{figure}[th]
    \centering
    \includegraphics[width=0.8\linewidth]{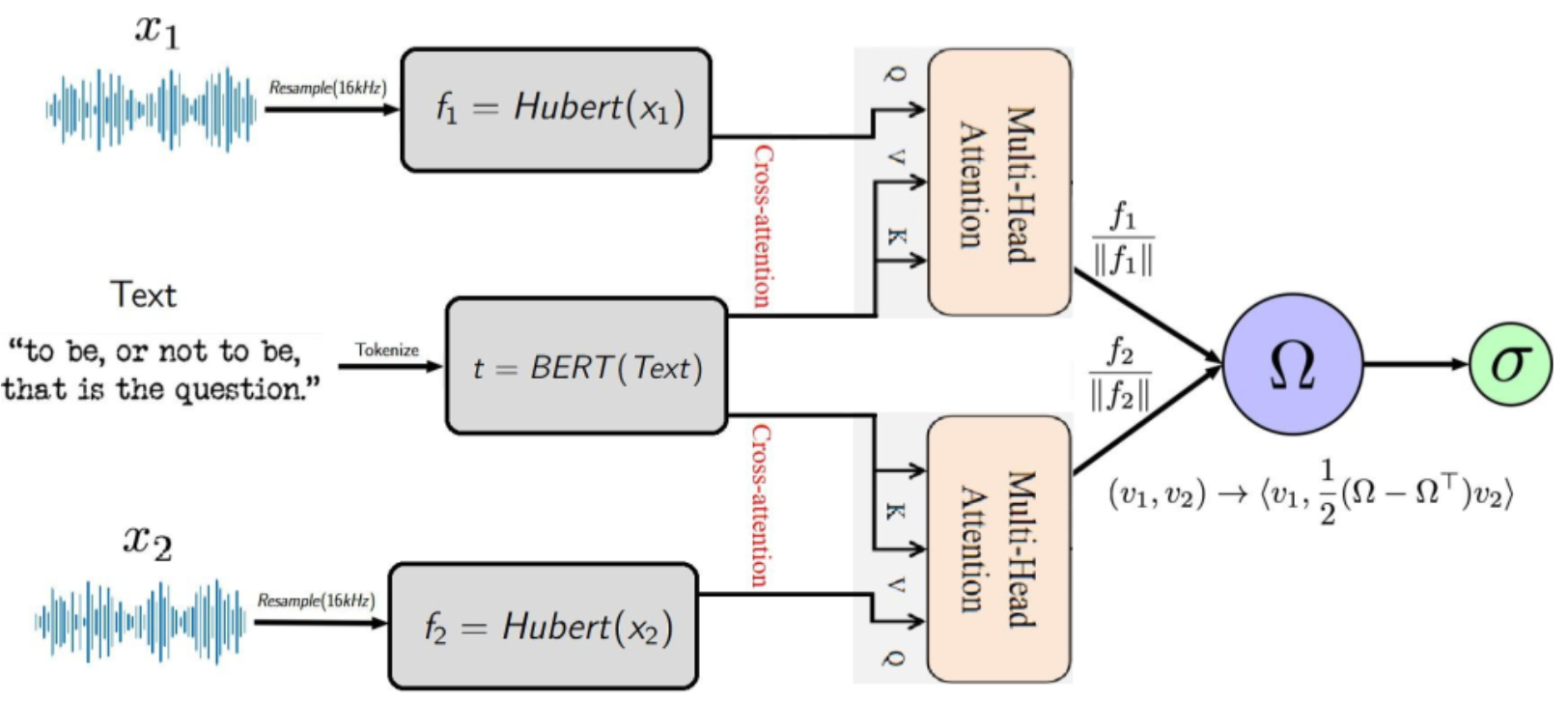}
    \caption{The NeuralSBSBert architecture. It extends NeuralSBS by incorporating textual features from BERT via a cross-attention mechanism, allowing the model to fuse audio and text information before the final preference decision.}
    \label{fig:neuralsbsbert}
\end{figure}

\textbf{NeuralSBSBert:} A multimodal variant (Figure~\ref{fig:neuralsbsbert}) that appends BERT text features via cross-attention, using audio embeddings as queries and text tokens as keys/values, attempting to bias the audio embedding toward content-aware cues (e.g., handling mispronunciations of the transcript).

\subsection{Architectures for absolute evaluation (MOS)}

While SBS determines preferences, absolute quality scoring is equally vital. We explored several directions for MOS prediction.

\textbf{Architectural baselines (CNNMOS, Attention-only):} To justify the choice of the CNN+BLSTM backbone, we additionally evaluated two simplified baselines. \textit{CNNMOS} removes the BLSTM component entirely and replaces it with deeper convolutional layers. \textit{Attention-only MOSNet} replaces the BLSTM with a self-attention pooling mechanism over audio features, without incorporating text. Both baselines achieved an RMSE of approximately 0.50, confirming the importance of the recurrent temporal modeling in the full MOSNet architecture.

\textbf{Enhanced MOSNet:} We utilized the classic MOSNet architecture (1D-CNNs followed by BLSTM) but introduced critical training enhancements adopted from NLP. Audio clips of varying lengths typically cause excessive padding in batches, distorting gradients. We implemented a custom DataLoader that sorts audio files by length prior to batching to minimize padding. Furthermore, we masked padding frames during training, excluding them from the final sequence-level average MOS calculation. We also replaced the original frame-level MSE loss with a sequence-level MSE, which yielded significantly more stable convergence. Additionally, we applied Dropout ($p=0.3$) after convolutional layers and Batch Normalization for regularization.

\begin{figure}[th]
    \centering
    \begin{minipage}[b]{0.48\linewidth}
        \centering
        \includegraphics[width=\linewidth]{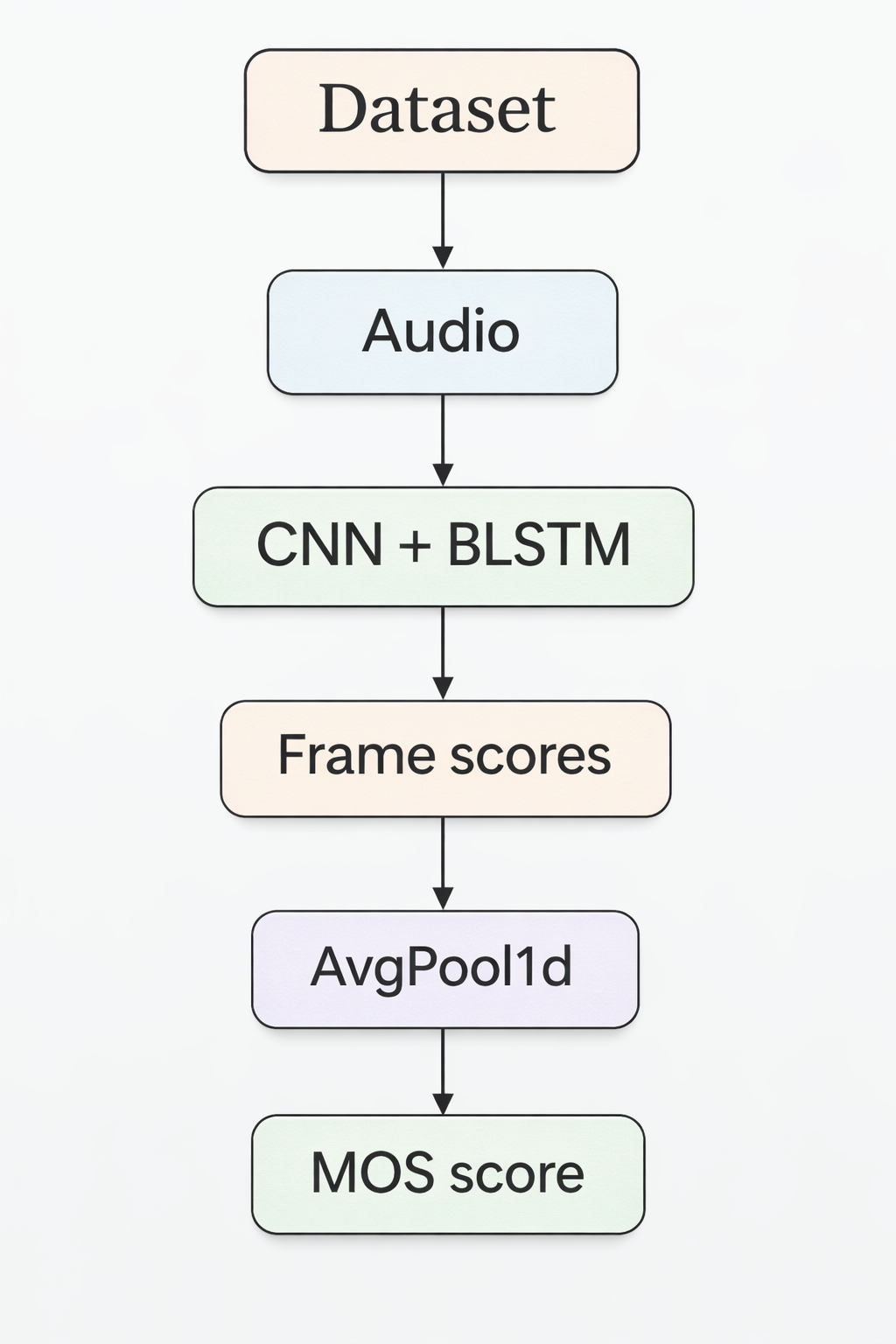}
        \caption{The enhanced MOSNet architecture: Audio features pass through 1D-CNNs and a BLSTM before global average pooling. Padding is explicitly masked during the loss calculation.}
        \label{fig:mosnet}
    \end{minipage}
    \hfill
    \begin{minipage}[b]{0.48\linewidth}
        \centering
        \includegraphics[width=\linewidth]{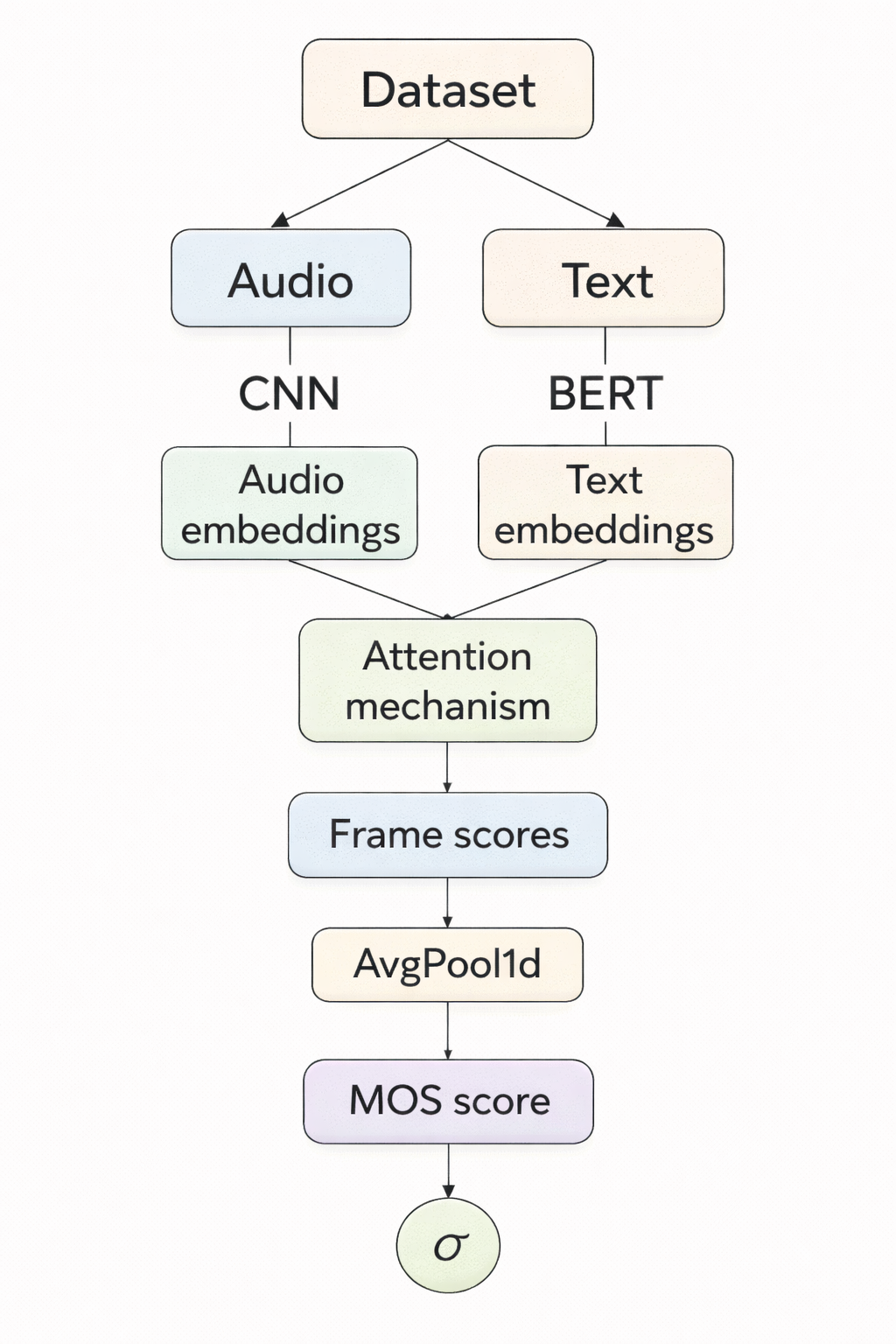}
        \caption{MOSNetBert architecture: It extends MOSNet by incorporating textual features from BERT via a cross-attention mechanism, allowing the model to fuse audio and text information before the final preference decision.}
        \label{fig:mosnetbert}
    \end{minipage}
\end{figure}

\textbf{MOSNetBert:} A multimodal variant (Figure~\ref{fig:mosnetbert}) that appends BERT text features via cross-attention, using audio embeddings as queries and text tokens as keys/values.

\begin{figure}[htbp]
    \centering
    \includegraphics[width=0.4\linewidth]{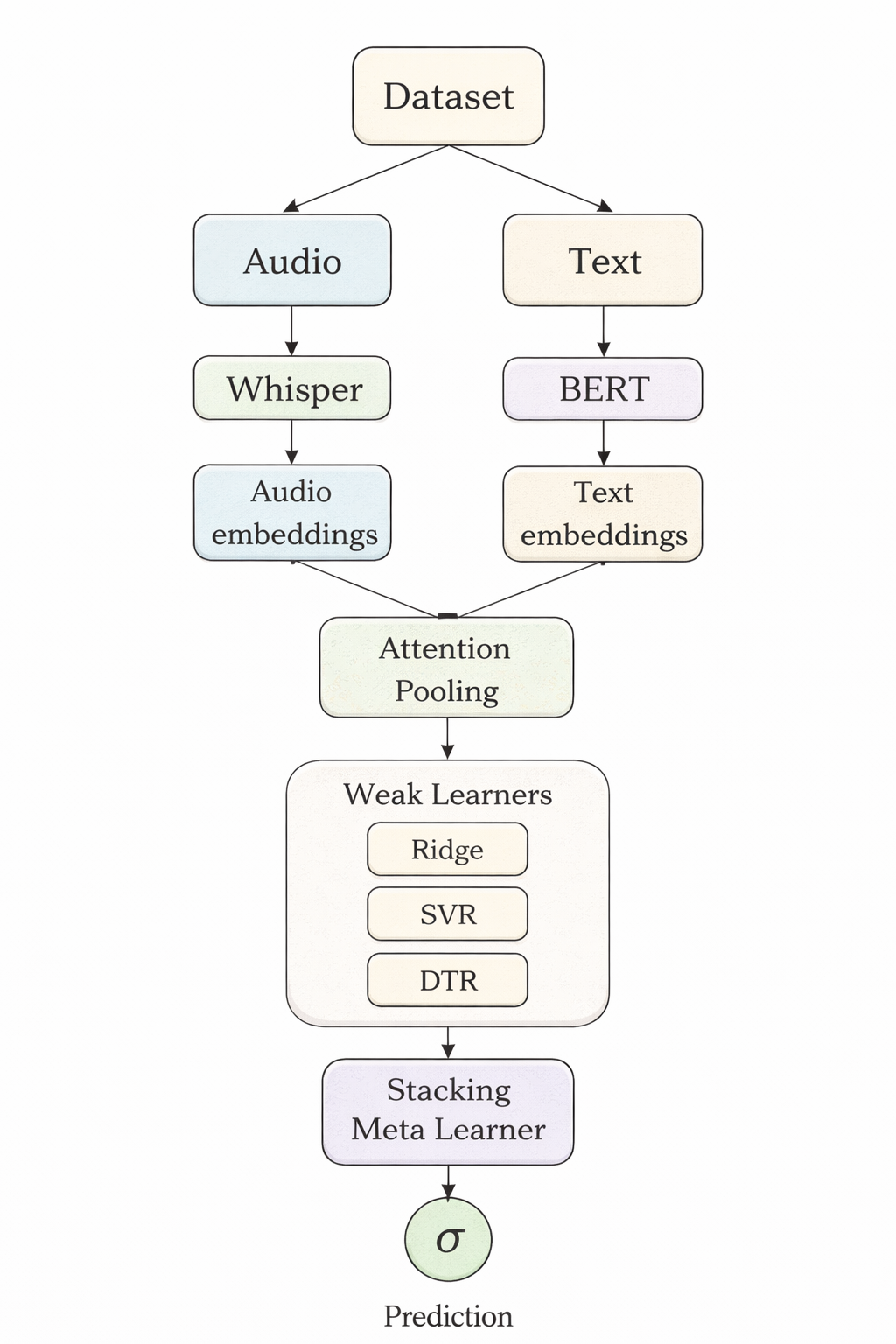} 
    \caption{The WhisperBert multimodal stacking architecture. Independent weak learners process concatenated audio (Whisper) and text (BERT) embeddings, supervised by a meta-learner.}
    \label{fig:whisperbert}
\end{figure}

\textbf{WhisperBert (Multimodal Stacking Ensemble):} Assuming that absolute quality perception relies on comparing acoustic delivery against expected textual semantics, we designed a stacking ensemble. The model extracts 512-dimensional temporal embeddings using OpenAI's \textbf{Whisper} (\textit{base.en}) via attention pooling, and global semantic embeddings from \textbf{BERT} (\textit{bert-base-uncased}) using the [CLS] token. Instead of direct cross-attention (which proved unstable in our SBS experiments), we concatenate these embeddings and pass them through a diverse set of weak learners (Ridge Regression, Support Vector Regression, and Decision Tree Regressor). The predictions of these weak learners are fed into a Meta-Learner (a 2-layer MLP with ReLU activation) to produce the final MOS score (Figure~\ref{fig:whisperbert}).

\textbf{SpeechLM-based architecture (SpeechLMMosTTS):} We also explored a more complex multimodal design built around a pretrained SpeechLM encoder for audio and BERT for text. Audio features from SpeechLM were passed through a three-layer convolutional extractor (with GELU activations and Batch Normalization), then fused with BERT embeddings via Multi-Head Attention. The fused representation was processed by a three-layer Transformer encoder and classified into five MOS categories through a series of gated residual blocks. Despite extensive hyperparameter search (233 training runs), the model consistently collapsed to predicting the dataset-wide mean MOS, achieving only 40\% classification accuracy. We attribute this failure to the difficulty of jointly optimizing a deep fusion module over a relatively small labeled dataset, reinforcing our preference for the simpler stacking approach of WhisperBert.

\section{Experiments and results}

\subsection{Relative evaluation (SBS) results}

On the SOMOS dataset, the audio-only NeuralSBS model attained 73.7\% accuracy and an AUC-ROC of 0.816 (Table \ref{tab:sbs-metrics}). Interestingly, the multimodal variant (NeuralSBSBert) slightly underperformed. This suggests that naive cross-attention fusion can disrupt robust audio cues. Furthermore, both models exhibit a noticeable performance drop on the standardized SOMOS dataset. This decline likely occurs because the standardization process, while intended to mitigate individual rater bias, compresses the overall variance of the scores, thereby narrowing the margins between paired samples and making relative quality distinctions harder to discriminate. This aligns with our error analysis, which reveals that misclassifications primarily occur when the absolute MOS difference between pairs is below 0.3 — a threshold where human raters themselves exhibit near-random agreement. Additionally, the model struggles to detect short-duration, low-intensity artifacts such as clicks that are not well captured in HuBERT embeddings.

\begin{table}[th]
  \caption{Performance of SBS models on original and standardized SOMOS datasets.}
  \label{tab:sbs-metrics}
  \centering
  \begin{tabular}{ l c c }
    \toprule
    \textbf{Metric} & \textbf{NeuralSBS} & \textbf{NeuralSBSBert} \\
    \midrule
    Accuracy (SOMOS)           & \textbf{0.737} & 0.727 \\
    AUC-ROC (SOMOS)            & \textbf{0.816} & 0.804 \\
    Accuracy (Std SOMOS)       & 0.717 & 0.712 \\
    AUC-ROC (Std SOMOS)        & 0.793 & 0.786 \\
    \bottomrule
  \end{tabular}
\end{table}

\subsection{Absolute evaluation (MOS) results}

The results for absolute MOS prediction highlight the substantial impact of our data standardization technique (Std SOMOS). As shown in Table \ref{tab:mos-metrics}, evaluating on the rater-standardized dataset consistently improved the Root Mean Square Error (RMSE) across all models. However, mirroring the trend observed in our pairwise evaluation, augmenting the baseline acoustic architecture with text features (MOSNetBert) actually degraded predictive accuracy compared to the vanilla MOSNet. This reaffirms that naive multimodal fusion can interfere with acoustic quality assessment, though the superior performance of WhisperBert suggests that text integration can be highly effective when paired with stronger, context-rich audio representations like Whisper.

\begin{table}[th]
  \caption{RMSE of absolute MOS predictors. Lower is better. Human inter-rater RMSE is $\approx 0.62$.}
  \label{tab:mos-metrics}
  \centering
  \resizebox{0.7\columnwidth}{!}{%
  \begin{tabular}{ l c c c }
    \toprule
    \textbf{Dataset} & \textbf{MOSNet} & \textbf{MOSNetBert} & \textbf{WhisperBert} \\
    \midrule
    RMSE (SOMOS)     & 0.491 & 0.547 & \textbf{0.457} \\
    RMSE (Std SOMOS) & 0.422 & 0.496 & \textbf{0.402} \\
    \bottomrule
  \end{tabular}%
  }
\end{table}

\subsubsection{MOSNet ablation study}

Table~\ref{tab:mosnet-ablation} details the incremental impact of each training enhancement introduced in our MOSNet pipeline. Starting from a vanilla configuration (no sorting, no masking, no regularization), we observe that Dropout and Batch Normalization reduce the RMSE from 0.531 to 0.491 on SOMOS. Length-based sorting yields an RMSE of 0.496, indicating improved gradient stability. Notably, an attempt to balance the score distribution via a weighted sampler degraded performance to 0.546, suggesting that the natural score distribution is important for calibration. The combination of all beneficial enhancements and the Std SOMOS targets yields the best result of 0.422.

\begin{table}[th]
  \caption{MOSNet ablation: incremental effect of training enhancements. All models share the same CNN+BLSTM backbone.}
  \label{tab:mosnet-ablation}
  \centering
  \begin{tabular}{ l c }
    \toprule
    \textbf{Configuration} & \textbf{RMSE} \\
    \midrule
    Vanilla (no sort, no mask, no reg.)       & 0.531 \\
    + Dropout (0.3) + BatchNorm               & 0.491 \\
    + Length sorting (SOMOS)                  & 0.496 \\
    + Weighted class sampler                  & 0.546 \\
    CNNMOS (no BLSTM, deeper CNN)             & $\sim$0.50 \\
    Attention-only (no BLSTM, no text)        & $\sim$0.50 \\
    \midrule
    Best config on Std SOMOS                  & \textbf{0.422} \\
    \bottomrule
  \end{tabular}
\end{table}

\subsubsection{WhisperBert: full metrics and ablations}

Table~\ref{tab:whisperbert-full} presents the complete set of evaluation metrics for WhisperBert on the original SOMOS and standardized Std SOMOS datasets. Data standardization substantially reduces MSE (from 0.209 to 0.161).

\begin{table}[th]
  \caption{Full evaluation metrics for WhisperBert on SOMOS and Std SOMOS.}
  \label{tab:whisperbert-full}
  \centering
  \begin{tabular}{ l c c }
    \toprule
    \textbf{Metric} & \textbf{SOMOS} & \textbf{Std SOMOS} \\
    \midrule
    MSE                       & 0.209 & \textbf{0.161} \\
    RMSE                      & 0.457 & \textbf{0.402} \\
    Linear Corr.\ (LCC)       & 0.599 & \textbf{0.565} \\
    Kendall's $\tau$          & 0.415 & \textbf{0.373} \\
    \bottomrule
  \end{tabular}
\end{table}

We conducted two additional ablation experiments on WhisperBert. First, we varied the Whisper encoder capacity by replacing \textit{base.en} with \textit{large} and \textit{turbo} variants; neither yielded an improvement in any metric, suggesting that the base model already captures sufficient acoustic information for MOS prediction. Second, removing the BERT text embeddings (audio-only WhisperBert) caused a visible degradation across all metrics, confirming that the textual modality contributes meaningfully when integrated via late-stage stacking rather than cross-attention.

\subsubsection{Comparison with UTMOS}

We benchmarked WhisperBert against UTMOS \cite{utmos}, a Wav2Vec 2.0-based MOS predictor widely used as a reference. Table~\ref{tab:utmos-comparison} reports the full set of system-level metrics. While UTMOS achieves substantially higher linear and rank correlations (LCC: 0.842, SRCC: 0.809), WhisperBert attains a lower absolute error (MSE: 0.161 vs.\ 0.242). This complementary profile suggests that UTMOS is better suited for system-level ranking, whereas WhisperBert excels at precise absolute score estimation --- a property more relevant for CI/CD gating scenarios where a fixed quality threshold must be evaluated.

\begin{table}[th]
  \caption{System-level comparison of WhisperBert (Std SOMOS) and UTMOS.}
  \label{tab:utmos-comparison}
  \centering
  \begin{tabular}{ l c c }
    \toprule
    \textbf{Metric} & \textbf{WhisperBert} & \textbf{UTMOS} \\
    \midrule
    MSE                  & \textbf{0.161} & 0.242 \\
    LCC                  & 0.565 & \textbf{0.842} \\
    SRCC                 & 0.373 & \textbf{0.809} \\
    Kendall's $\tau$     & 0.373 & \textbf{0.625} \\
    \bottomrule
  \end{tabular}
\end{table}

\subsection{LLM-based evaluation experiments}

To assess the viability of general-purpose LLMs as TTS evaluators, we tested two systems in zero-shot settings. For this experiment, we constructed a controlled evaluation subset consisting of 1,000 clean audio clips, where each clean clip was paired with an augmented version containing specific induced distortions. On this relatively straightforward discrimination task, our dedicated models (NeuralSBS and WhisperBERT) achieved a near-perfect accuracy of 99.8\%.

\textbf{Qwen2-Audio.} We evaluated Qwen2-Audio-7B-Instruct for absolute MOS scoring and Qwen-Audio-Chat (which supports dual audio inputs) for SBS comparison. The model received a structured prompt describing the evaluation criteria (naturalness, intonation, absence of distortions) and was asked to output a numeric score. MOS accuracy reached only 31.25\%, while SBS accuracy was 76.00\%. Given the clear contrast between clean and distorted audio in this subset, this 76.00\% accuracy falls drastically short of our 99.8\% baseline, indicating limited sensitivity to even explicit acoustic artifacts.

\textbf{Gemini 2.5 Flash Preview.} We additionally experimented with Gemini 2.5 Flash Preview for data annotation. The model performed poorly on absolute MOS regression but showed acceptable SBS discrimination when listening to audio pairs. At a cost of approximately 0.15\$ per 1M input tokens (3.50\$/1M output), each SBS comparison consumed roughly 1,400 tokens. While potentially useful for bootstrapping SBS labels on new languages where human annotations are scarce, the model's MOS predictions were insufficiently calibrated for training data generation.

These results underscore that, despite the rapid progress of multimodal LLMs, dedicated regression architectures with metric-specific inductive biases remain necessary for reliable TTS evaluation, even on elementary degradation tasks.

\section{Conclusion}

This work demonstrates that automated, neural evaluation of TTS quality can surpass individual expert-level subjective assessment reliability while dramatically reducing cost and latency. By addressing both relative (SBS) and absolute (MOS) evaluation paradigms, we draw several critical conclusions:

\begin{itemize}
    \item \textbf{Human-level reliability:} NeuralSBS correctly identifies the human-preferred audio in nearly 74\% of cases, a rate commensurate with human subjective agreement bounds. Furthermore, absolute predictors like WhisperBert and MOSNet achieve RMSEs ($\approx 0.40$) well below the human inter-rater baseline ($\approx 0.62$).
    \item \textbf{The importance of data standardization:} Mitigating rater bias via collaborative filtering techniques (Std SOMOS) significantly and consistently improves the learning stability and final RMSE of absolute MOS predictors.
    \item \textbf{Challenges of multimodality:} While adding textual transcripts intuitively seems beneficial, direct latent fusion via Cross-Attention (as seen in NeuralSBSBert and MOSNetBert) often degrades performance. The SpeechLM-based architecture, despite 233 hyperparameter search runs, collapsed to predicting the dataset mean. Conversely, late-stage fusion via ensemble stacking (WhisperBert) successfully leverages multimodal signals to improve absolute accuracy.
    \item \textbf{Negative results with LLMs:} Zero-shot evaluation with Qwen2-Audio and Gemini 2.5 Flash Preview yields substantially inferior accuracy compared to specialized architectures, reinforcing the need for dedicated metric learning.
\end{itemize}

Practically, these evaluators can be seamlessly embedded in CI/CD pipelines to gate TTS releases, enabling automated, reproducible quality control at scale. Future work will focus on full cross-lingual evaluation (particularly Russian), integration of visual modalities for video-TTS assessment, and self-supervised pretraining strategies that reduce dependence on human annotations.

\bibliographystyle{IEEEtran}
\bibliography{mybib} 

@article{speechgpt,
  title={{SpeechGPT}: Empowering Large Language Models with Intrinsic Cross-Modal Conversational Abilities},
  author={Zhang, Y. and others},
  journal={arXiv preprint arXiv:2305.11000},
  year={2023}
}

@inproceedings{somos,
  title={{SOMOS}: The Samsung Open {MOS} Dataset for the Prediction of Synthesized Speech Quality},
  author={Chessex, G. and others},
  booktitle={Proc. Interspeech},
  year={2022}
}

@misc{voicemos,
  title={{VoiceMOS} Challenge 2024},
  author={{VoiceMOS Challenge}},
  howpublished={\url{https://sites.google.com/view/voicemos-challenge}},
  year={2024}
}

@article{mosnet,
  title={{MOSNet}: Deep Learning based Objective Assessment for Voice Conversion},
  author={Lo, C.-C. and others},
  journal={arXiv preprint arXiv:1904.08352},
  year={2019}
}

@inproceedings{nsbs,
  title={Neural Side-by-Side: Predicting Human Preferences for No-Reference Super-Resolution Evaluation},
  author={Khrulkov, V. and Babenko, A.},
  booktitle={Proceedings of the IEEE/CVF Conference on Computer Vision and Pattern Recognition (CVPR)},
  year={2021}
}

@article{hubert,
  title={{HuBERT}: Self-Supervised Speech Representation Learning by Masked Prediction of Hidden Units},
  author={Hsu, W.-N. and Bolte, B. and Tsai, Y.-H. H. and Lakhotia, K. and Salakhutdinov, R. and Mohamed, A.},
  journal={arXiv preprint arXiv:2106.07447},
  year={2021}
}

@inproceedings{dnsmos,
  title={{DNSMOS}: A Non-Intrusive Perceptual Objective Speech Quality Metric to Evaluate Noise Suppressors},
  author={Reddy, C. K. A. and others},
  booktitle={Proc. ICASSP},
  pages={6493--6497},
  year={2021}
}

@article{distilmos,
  title={{DistilMOS}: Layer-Wise Self-Distillation For Self-Supervised Learning Model-Based {MOS} Prediction},
  author={Yang, Jianing and Nakata, Wataru and Saito, Yuki and Saruwatari, Hiroshi},
  journal={arXiv preprint arXiv:2601.13700},
  year={2026}
}

@article{nisqa,
  title={{NISQA}: A Deep {CNN}-Self-Attention Model for Multidimensional Speech Quality Prediction},
  author={Mittag, G. and M{\"o}ller, S.},
  journal={Proc. Interspeech},
  pages={2127--2131},
  year={2021}
}

@inproceedings{lpips,
  title={The Unreasonable Effectiveness of Deep Features as a Perceptual Metric},
  author={Zhang, R. and others},
  booktitle={Proc. CVPR},
  pages={586--595},
  year={2018}
}

@inproceedings{utmos,
  title={{UTMOS}: {UTokyo-SaruLab} System for {VoiceMOS} Challenge 2022},
  author={Saeki, Takaaki and Xin, Detai and Nakata, Wataru and Koriyama, Tomoki and Takamichi, Shinnosuke and Saruwatari, Hiroshi},
  booktitle={Proc. Interspeech},
  pages={4546--4550},
  year={2022}
}

\end{document}